\documentclass{article}

\usepackage{microtype}
\usepackage{graphicx}
\usepackage{subcaption}
\usepackage{multirow}
\usepackage{booktabs} % for professional tables
\usepackage{listing}

\usepackage{hyperref}

\usepackage[preprint]{icml2026}

\usepackage{amsmath}
\usepackage{amssymb}
\usepackage{mathtools}
\usepackage{amsthm}

\usepackage[capitalize,noabbrev]{cleveref}

\theoremstyle{plain}

\theoremstyle{definition}

\theoremstyle{remark}

\usepackage[textsize=tiny]{todonotes}

\icmltitlerunning{Formalizing Numerical Analysis: An Agent Pipeline and Quality Audit Beyond Kernel Acceptance}

\begin{document}

\twocolumn[
  \icmltitle{Formalizing Numerical Analysis: An Agent Pipeline and Quality Audit Beyond Kernel Acceptance}



  \icmlsetsymbol{equal}{*}

  \begin{icmlauthorlist}
    \icmlauthor{Theodore Meek}{equal,mathai,uwmath}
    \icmlauthor{Siyuan Ge}{equal,mathai,uwcs}
    \icmlauthor{Di Qiu Xiang}{mathai,uwmath,uwcs}
    \icmlauthor{Simon Chess}{mathai,uwmath,uwcs}
    \icmlauthor{Vasily Ilin}{mathai,uwmath,uwamath}
  \end{icmlauthorlist}
  
  \icmlaffiliation{mathai}{Math AI Lab, University of Washington, Seattle, United States}
  \icmlaffiliation{uwmath}{Department of Mathematics, University of Washington, University of Washington, Seattle, United States}
  \icmlaffiliation{uwamath}{Department of Applied Mathematics, University of Washington, Seattle, United States}
  \icmlaffiliation{uwcs}{Department of Computer Science \& Engineering, University of Washington, Seattle, United States}
  %\icmlaffiliation{comp}{Company Name, Location, Country}

  \icmlcorrespondingauthor{Theodore Meek}{theomeek@uw.edu}
  \icmlcorrespondingauthor{Vasily Ilin}{vilin@uw.edu}

  \icmlkeywords{Machine Learning, ICML}

  \vskip 0.3in]



\printAffiliationsAndNotice{\icmlEqualContribution}  % no special notice (required even if empty)

\begin{abstract}
Recent work has demonstrated that coding agents can formalize entire advanced mathematics textbooks in Lean 4, yet existing efforts concentrate on branches of mathematics already well-represented in \texttt{mathlib} and measure success solely through kernel acceptance. We address both limitations by applying a coding agent to formalize \textit{Numerical Methods for Ordinary Differential Equations}, a textbook in numerical analysis that is largely absent from \texttt{mathlib}, stressing the agent's capacity to develop new theory from scratch. We further introduce a systematic, reproducible three-dimensional framework for evaluating the quality of agent-produced formalizations beyond compilation: semantic correctness, Mathlib reuse, and cross-file reuse via LLM-as-judge methods. Applying this framework to our own formalization and to the released outputs of RepoProver and M2F, we uncover recurring unfaithful formalization patterns---including incomplete multi-part statements, added weakening hypotheses, and parameter restrictions---that kernel acceptance entirely obscures. Our results suggest that compilation-based metrics substantially overstate formalization quality, and we provide a reproducible audit methodology to support more rigorous evaluation of future autoformalization systems.
\end{abstract}

\section{Introduction}

Interactive theorem provers such as Lean~\cite{demoura2021lean} make it possible to verify mathematics with machine-checked rigor. Lean's mathematical library, \texttt{mathlib}~\cite{The_mathlib_Community_2020}, now contains hundreds of thousands of definitions and theorems spanning much of the undergraduate and graduate curriculum, making Lean an increasingly practical foundation for verifying research-level mathematics and for building machine-checkable mathematical knowledge at scale.

The principal obstacle to formalization at scale is its cost. Translating mathematics written in natural language into formal code is slow and labor-intensive and demands expertise in both mathematics and formal language. This obstacle has motivated a substantial body of work on autoformalization, attempts to use language models to automatically translate informal mathematics into formal statements and proofs~\cite {szegedy2020promising, wu2022autoformalizationlargelanguagemodels, jiang2023draft}. 

Three families of approaches have emerged. Specialized models trained specifically for formal mathematics have shown their performance on per-problem benchmarks such as miniF2F~\cite{zheng2022minif2f}, ProofNet~\cite{azerbayev2023proofnet}, and PutnamBench~\cite{tsoukalas2024putnambench}, with recent systems including DeepSeek-Prover-V2~\cite{ren2025deepseekproverv2}, Goedel-Prover-V2~\cite{lin2025goedelproverv2}, Kimina-Prover~\cite{wang2025kiminaprover}, and Aristotle~\cite{achim2025aristotle}. General-purpose frontier models, applied through prompting and multi-model cooperation, have proven surprisingly capable at formalization without task-specific training~\cite{wu2022autoformalizationlargelanguagemodels}. Most recently, coding agents, which are frontier models equipped with tools, file access, compiler feedback, and the ability to act over many steps, have been used to formalize mathematics at the scale of entire textbooks, producing formal developments of hundreds of pages in days~\cite{urban2026topology, m2f2026, repoprover2025, liu2026numinaleanagent, ren2026merlean}.

As autoformalization moves from isolated problems to whole textbooks, two concerns become pressing. The first is generalizability. Most existing results concern mathematics already well represented in \texttt{mathlib}, and it is far less clear how these systems behave on mathematics that \texttt{mathlib} does not yet cover, where the agent must develop the relevant theory from the ground up rather than retrieve it~\cite{kumarappan2025leanagent, repoprover2025}. The second is evaluation. Existing whole-textbook formalization efforts report ``success'' almost exclusively as whether the code type-checks and builds without \texttt{sorry}. However, kernel acceptance establishes only that a formalization is well-formed and compiled, not that it is semantically correct~\cite{poiroux2024typechecking, li2024symbolic, liu2025rethinking, lu2025formalalign, weng2025survey}. For example, a compiled statement may be vacuously true~\cite{li2024symbolic} or may misinterpret quantifier scopes or implicit constraints~\cite{zhang2025beyondgold, chen2025reform}. Yet in the whole-textbook setting, quality beyond compilation is assessed only through limited, subjective spot-checks~\cite{m2f2026, repoprover2025}.

In this work, we address both concerns. First, we apply a coding agent to formalize \emph{Numerical Methods for Ordinary Differential Equations}, a textbook on the branch of numerical analysis that is almost entirely absent from \texttt{mathlib}, stressing the agent's ability to develop new theory rather than recombine existing library content. Second, we introduce a systematic and reproducible method for evaluating the quality of the resulting formalization, auditing it along three dimensions: (1)~\emph{semantic correctness}: whether each formalized statement faithfully captures the source; (2)~\emph{mathlib reuse}: whether the agent reuses existing \texttt{mathlib} results when it should, rather than re-formalizing them; and (3)~\emph{cross-file reuse}: whether it reuses results already formalized elsewhere in the project, rather than duplicating them. Our method combines automated analysis of the formalized code with an LLM-as-judge~\cite{zheng2023judging, zhang2025beyondgold}, and we demonstrate it through a case study on our own formalization and RepoProver and M2F’s released results.

Our contributions are:
\begin{itemize}
    \item We apply a coding agent to formalize a textbook in numerical analysis, a branch of mathematics largely uncovered by \texttt{mathlib}, providing a stress test of agentic autoformalization outside the well-covered domains studied in prior work.
    \item We introduce a systematic, reproducible method that evaluates formalization quality beyond kernel acceptance along three dimensions---semantic correctness, mathlib reuse, and cross-file reuse---using LLM-as-judge.
    \item We report a detailed evaluation of our formalization under this method, characterizing where and how current agentic formalization succeeds and falls short.
\end{itemize}

\section{Related Work}

\subsection{Agentic Autoformalization}
\label{subsec:agentic}

Early autoformalization treated translation as a single-shot, per-statement task~\cite{wu2022autoformalizationlargelanguagemodels, jiang2023draft, azerbayev2023proofnet}, and specialized provers continue to push statement- and proof-level performance on benchmarks such as miniF2F and ProofNet~\cite{ren2025deepseekproverv2, lin2025goedelproverv2, wang2025kiminaprover}. A more recent line equips frontier models with tools, file access, compiler feedback, and multi-step control, yielding \emph{agentic} formalization systems~\cite{liu2026numinaleanagent, ren2026merlean}. Several concurrent efforts have scaled this paradigm to entire textbooks and projects: a recent effort formalized general topology in Megalodon~\cite{urban2026topology}; M2F compiled hundreds of pages of analysis into a buildable Lean project~\cite{m2f2026}; RepoProver orchestrated thousands of parallel agents to formalize a graduate algebraic-combinatorics textbook in Lean~\cite{repoprover2025}; and further multi-agent and cross-system efforts target additional textbook material~\cite{brown2026agenthunt, bryant2026munkres}. These works demonstrate that agentic formalization at the textbook scale is feasible, but they work on a well-formalized branch of mathematics and report success primarily as kernel acceptance. In contrast, we study a branch largely absent from \texttt{mathlib}, and we focus on evaluating the quality of agent-produced formalization.

\subsection{Formalization Evaluation}

It is well-established that compilation in a proof assistant is a necessary but not sufficient condition for a faithful formalization~\cite{poiroux2024typechecking, li2024symbolic, liu2025rethinking, weng2025survey}. Surface-form metrics such as BLEU correlate poorly with correctness~\cite{lu2025formalalign}, motivating dedicated approaches: symbolic and bidirectional-equivalence checking~\cite{li2024symbolic, poiroux2024typechecking}, back-translation and learned alignment models~\cite{lu2025formalalign, gao2025herald}, and LLM-as-judge methods~\cite{peng2025criticlean, zhang2025beyondgold, chen2025reform}. These methods, however, evaluate isolated statements against a single informal counterpart. Our work differs in two aspects: we evaluate agent-produced formalization at the project scale rather than per statement, and we introduce structural dimensions by analyzing mathlib usage and cross-file usage, which assess how a formalization integrates with the library and project. Such an integration becomes essential for large-scale formalization.

\section{Methods}

\subsection{Formalization Pipeline}

Our pipeline is an autonomous loop driven by a Python
orchestrator that repeatedly invokes four LLM roles: a
\textit{Planner}, a \textit{Worker}, an \textit{Evaluator}, and a
\textit{Consultant} summoned only on escalation. Like recent
project-scale autoformalization frameworks such as
RepoProver \cite{repoprover2025} and M2F \cite{m2f2026}, we target
whole Lean projects rather than isolated theorems. We only commit edits when Lean builds. We differ from M2F's fixed two-stage statement then proof pipeline by running an open-ended loop whose target is chosen each cycle from the live repository state. The Planner and Worker roles are implemented using Claude Opus 4.6; the Evaluator uses Claude Sonnet. We replace RepoProver's sketcher/prover/reviewer split with a Planner/Worker/Evaluator/Consultant decomposition coordinated through GitHub Actions.

\paragraph*{Stateless agents and shared state.}
The agents themselves are stateless: no agent retains memory across
cycles. All inter-cycle communication happens through files committed
to the repository and through GitHub itself. The key artifacts are
\texttt{strategy.md}, written by the Planner and read by the Worker
(overwritten each cycle); \texttt{task\_results/cycle\_NNN.md}, the
Worker's per-cycle report; \texttt{issues/$\langle$name$\rangle$.md},
blocker descriptions; \texttt{history.jsonl}, one record per cycle
holding the Evaluator's structured judgment (score, summary,
stuck-on, recommendation); and \texttt{cycle},
\texttt{heartbeat.json}, and \texttt{run.lock}, which hold control
state for the orchestrator and watchdog.

\paragraph*{Per-cycle pipeline.}
Each cycle proceeds through the following stages:
\begin{itemize}
  \item \textbf{Synchronization.} The orchestrator pulls the latest
        commits and queries GitHub Actions for build status. If the
        Lean build is failing, the Planner is hard-coded to fix the
        build first.
  \item \textbf{Planning.} The Planner reads \texttt{plan.md}, current
        \texttt{sorry} locations, recent task results, open issues,
        and recent cycle history, and emits a concrete instruction
        set in \texttt{strategy.md}.
  \item \textbf{Execution.} The Worker edits the Lean source tree
        following a sorry-first strategy: state the theorem in full,
        verify that the skeleton compiles, then close each
        \texttt{sorry} one at a time. It uses Lean LSP tools and
        Mathlib search (LeanSearch \cite{leansearchv2},
        Loogle \cite{loogle}). It commits its changes and writes a
        task-results file; if blocked, it opens an issues file.
  \item \textbf{Evaluation.} A separate LLM session reads the cycle's
        Git diff, the Worker's task result, and recent attempts, and
        emits a progress score in $[-2, 2]$, a one-sentence summary,
        the current blocker, a recommendation for the next Planner
        cycle, and a compacted line for \texttt{attempts.md}.
  \item \textbf{Mechanical gates.} Independently of the LLM judge,
        the orchestrator enforces hard constraints. We adopt an
        M2F-style verifier-certified refinement \cite{m2f2026}
        check on the total \texttt{sorry} count across the source
        tree, forbidding it from strictly increasing; budget
        increases are rate-limited and require a rationale. Strategy
        compliance is enforced by regex-extracting the scope declared
        in \texttt{strategy.md} and checking the Worker's commits
        against it. A build gate ensures that modified files compile
        before the Worker is permitted to commit.
  \item \textbf{Escalation.} If recent history shows four or more
        consecutive non-positive scores, the Consultant role is
        invoked, and its advice is written to the issue tracker for
        the next Planner cycle.
  \item \textbf{Bookkeeping.} The evaluation record is appended to
        \texttt{history.jsonl} and the cycle counter is advanced.
\end{itemize}

\paragraph*{Agent skills.}
Each agent is shaped by repository-level instructions (a
\texttt{CLAUDE.md} or \texttt{AGENTS.md} file) automatically loaded at
session start. These standing rules absorb content that, in earlier
iterations, had to be supplied through repeated human prompting.

\subsection{Analysis Methods}

\subsubsection{Semantic Correctness}
\label{sec:semantic-correctness}

We audit semantic faithfulness with two LLM-judge protocols that
share a rubric but differ in what the judge reads.

\paragraph*{Per-project NL/Lean pairing.}
Each corpus tracks formalized entities differently, so we extract
per-entity records pairing a textbook NL statement with its Lean
declaration header three ways. For Butcher we consume a curated
\texttt{lean\_status.json} with extracted NL. For the
alg-comb / RepoProver corpus we walk the blueprint
(\texttt{blueprint/src/chapter\_*.tex}), whose environments carry
\texttt{\textbackslash lean\{...\}} and
\texttt{\textbackslash leanok} macros that we parse directly. For
ReasBook, which ships only Lean, we pair each M2F-generated docstring
(\texttt{/-- Theorem N.M ... -/}) with the following declaration; the
NL is M2F's own gloss for $1{,}834$ of $1{,}864$ entities, so direct
judgment on ReasBook is largely a self-consistency check, with the
$30$ Rockafellar~\S1.1 entities the only true-fidelity samples. In every case we keep only the Lean theorem statement and discard the proof.
\paragraph*{Shared rubric.}
Each protocol compares two statements $A$ and $B$ by first answering
$Q_1\colon A \Rightarrow B$ and $Q_2\colon B \Rightarrow A$. The score
follows from a fixed truth table: $Q_1 \land Q_2$ yields \textbf{Faithful}
(verbatim); $Q_1 \land \neg Q_2$ yields \textbf{Stronger} ($A$ strictly
stronger, a sound generalization, acceptable for a formalization
audit); $\neg Q_1$ yields \textbf{Different} (divergent), except when multiple
concatenated Lean declarations together cover $A$ as a case split. We
collapse $\{\text{Stronger, Faithful}\}$ to be considered reasonable formalizations and $\{\text{Different}\}$ to an unreasonable formalization. The rubric also demands semantic similarity in addition to the implication requirements detailed above. It also handles when extra Lean hypotheses make the statement \emph{weaker}, not stronger, and will correct the score to Different.
\paragraph*{Direct judgment (Lean vs.\ NL).}
The judge sees the textbook NL, the Lean declaration header, and any
surrounding textbook prose, which may carry implicit hypotheses; $A$
is the Lean statement, $B$ the NL. We use DeepSeek-V4-Pro \cite{deepseekv4}.
\paragraph*{Round-trip judgment (NL vs.\ back-translated NL).}
A blind back-translator sees only the Lean and produces an NL
rendering. The judge then compares textbook NL against that
back-translation, following \cite{wu2022autoformalizationlargelanguagemodels}, who observe that informalization is empirically easier than formalization.
The dual-judge design follows \cite{li2024autoformalize}. Here $A$
is the back-translation and $B$ the NL, with $Q_1$ primary: failure
of $Q_1$ means the Lean cannot recover the textbook claim.
Back-translation uses DeepSeek V4 Pro; the round-trip judgment uses
\texttt{gpt-oss-120b} \cite{openai2025gptoss}, a deliberately
different model family to avoid correlated blind spots.

\subsubsection{Mathlib Usage}
We evaluate the formalizations on the originality of their contribution. As a heuristic for originality, we consider a proved statement (a \texttt{theorem}, \texttt{lemma}, etc) to be unoriginal if it can be proved by Mathlib theorems using \texttt{exact?+all}, a tactic which searches the available context for very simple proofs relying on a single theorem (if the theorem cannot be proven by anything in Mathlib we accept it as original, even if it is provable using \texttt{exact} in conjunction with a theorem proved by the project itself). In this way, we detect the extent to which a project is reproducing known results. It is worth noting that it is potentially healthy for a project to have a small number of theorems that are not considered ``original'' in this way. These may occur due to stylistic choices favoring shorter proofs. Nevertheless, a project that makes primarily novel contributions should consist of theorems primarily not provable by \texttt{exact?} under Mathlib.

\subsubsection{Crossfile Usage}
\label{crossfile_methods}

A high-quality formalization should reuse the results it has already established: when a textbook result $S$ depends on an earlier result $T$, the Lean proof of $S$ should invoke the formalized $T$ rather than re-deriving it or routing around it. We measure this form of internal reuse by checking whether each textbook-extracted dependency is reflected as a reference in the Lean code. The analysis has two stages: we first extract a dependency graph from the textbook, and then test whether each dependency is reflected in the formalization.

\paragraph{Textbook dependency graph.}
We segment each textbook into entities (i.e., definitions, theorems, lemmas, propositions, and corollaries, organized as blocks in the textbook) and pair each statement with its proof. The pipeline accepts both unstructured input (text extracted from a PDF) and a structured \LaTeX{} blueprint, from which entity identifiers and the associated Lean symbols are read directly from \verb|\label{}| and \verb|\lean{}| macros. Over these entities, we generate candidate dependency edges of $S use T$, using two heuristics.
\begin{itemize}
    \item \textbf{Citation edges} are extracted when $S$ explicitly cites $T$'s index. The index is a textbook number in the block title, such as Theorem~142C’’  or a \verb|\ref| to $T$'s \verb|\label| in the blueprint.
    \item \textbf{Keyword edges} are extracted when $S$ mentions a term that $T$ introduces. The introduced terms include the $T$'s name in its block title, a quoted phrase in $T$'s statement, and $T$'s \verb|\emph{}| term. 
\end{itemize}
Both heuristics generate false-positive edges, and keyword edges in particular fire on generic vocabulary and cross-domain homonyms under different qualifiers. We therefore apply an LLM filtering: for each candidate edge, we prompt \texttt{deepseek-v4-pro} to verify the dependency based on the textbook content of both endpoints and the evidence that triggered the edge. The model returns a strict JSON verdict with a confidence and rationale, and we retain only edges judged real.

\paragraph{Reflection in the Lean code.}
For each textbook-extracted dependency $S \to T$, we obtain the Lean declaration of both endpoints and test whether $S$ names $T$ within its declaration body (statement and proof term). Because the test matches names verbatim, an implicit dependency via an intermediate helper lemma, a \texttt{simp} set, or definitional unfolding will not be registered. Therefore, the reported rate is a lower bound on true code-level dependence.

\section{Results}

\subsection{Formalization Results}

We applied the pipeline to the Butcher textbook on
numerical methods for ordinary differential
equations \cite{butcher2016numerical}. Of the $175$ statements in the
textbook, the pipeline produced fully proof-complete Lean
formalizations for $84$ ($48.0\%$), and an additional $12$ ($6.9\%$)
with the Lean statement in place and at least one \texttt{sorry}
remaining, for a combined coverage of $54.9\%$. Coverage is uneven
across chapters: four of the
five chapters exceed $74\%$ combined formalization, while Chapter~3
on Runge--Kutta methods, the largest by a wide margin and
covering rooted trees, order conditions, A-stability, Pad\'e
approximants, order stars, AN/BN/algebraic stability, and the
Butcher group, sits at $34.8\%$ and accounts for the bulk of
unformalized material.

\subsubsection{Proof Gap}
 Proof completeness and axiom usage results are reported in Table~\ref{tab:sorry-axiom-comparison}. RepoProver's 5 open sorries correspond entirely to intentional exercise placeholders. The methods in which these are obtained are in Appendix~\ref{app:sorry-axiom-counting}
 \begin{table}[t]
\centering
\caption{Number of open sorries and unnecessary axioms across formalization systems. $^*$All sorries in RepoProver correspond to exercises and are intentional.}
\vskip 0.1in
\small
\begin{tabular}{lcc}
\toprule
\textbf{System} & \textbf{Sorries} & \textbf{Axioms} \\
\midrule
RepoProver & 5$^*$  & 0 \\
M2F        & 12     & 4 \\
OpenMath   & 0      & 0 \\
\bottomrule
\end{tabular}
\label{tab:sorry-axiom-comparison}
\vskip 0.1in
\end{table}

\subsection{Semantic Correctness}
Results are shown in Table~\ref{tab:semantic-correctness}. Agreement percentage is found by comparing results from  NL vs. Lean and NL vs. back-
translated NL after grouping faithful and stronger categories together.

\begin{table*}[t]
\centering
\caption{Semantic correctness evaluation across formalization systems via NL vs.\ Lean (direct, DeepSeek) and NL vs.\ back-translated NL (round-trip, GPT-oss-120b). Faithful = formalization is logically equivalent to source; Stronger = formalization entails source; Different = formalization diverges from source. $^\dagger$Agreement is computed by grouping faithful and stronger together.}
\label{tab:semantic-correctness}
\vskip 0.1in
\small
\begin{tabular}{llccc}
\toprule
& & \textbf{OpenMath} & \textbf{M2F} & \textbf{RepoProver} \\
\midrule
\scriptsize NL vs.\ Lean
  & Faithful  & 54\% (51)     & 73\% (1359) & 82\% (1173) \\
  & Stronger  & 4\% (4)       & 2\% (42)    & 1\% (12)    \\
  & Different & 42\% (40)     & 25\% (463)  & 18\% (253)  \\
\midrule
\scriptsize NL vs.\ NL$_{\text{b}}$
  & Faithful  & 46\% (44)     & 61\% (1050) & 69\% (931)  \\
  & Stronger  & 14\% (13)     & 13\% (225)  & 11\% (152)  \\
  & Different & 40\% (38)     & 26\% (447)  & 19\% (258)  \\
\midrule
\multicolumn{2}{l}{Agreement$^\dagger$} & 72.6\%  & 74.7\%      & 78.4\%      \\
\bottomrule
\end{tabular}
\vskip 0.1in
\end{table*}

\subsubsection{Qualitative results}
\label{sec:semantic-qualitative}

To validate the direct judge, two of the authors with Lean expertise
drew a stratified random sample of $20$ OpenMath entities, $10$
judged faithful or stronger (\textit{OK}) and
$10$ judged divergent (\textit{DIV}) with seed $42$, and
re-graded each one independently using the same $Q_1/Q_2$ rubric the
judge was given. Disagreements were resolved by discussion.

We agreed with the judge on $10/10$ OK verdicts and $8/10$ DIV
verdicts, for $18/20$ overall. The eight agreed-DIV cases were divergences from the textbook, falling into recurring
patterns documented below. The two disagreements both went in the
same direction; the judge marked DIV entities we considered
faithful, both times, because the LLM didn't understand standard Mathlib convention.
Additionally, we inspected the four entities marked Stronger by the
judge, and found that two are sound generalizations
exhibiting interesting strengthenings.

\paragraph*{Divergence: incompleteness.}
The most common DIV pattern in our sample is the Lean formalization
covering only part of a multi-part textbook statement. Definition
312A defines three quantities simultaneously, elementary weights
$\Phi(t)$, internal weights $\Phi_i(t)$, and derivative weights
$(\Phi_i D)(t)$, through four coupled equations; the Lean
formalization defines only \texttt{derivativeWeight}. Similarly,
Definition 356B defines both DJ-reducibility and the explicit
construction of the reduced method; the Lean defines only the
reducibility predicate. These cases would require the case-split
exception to recover, which applies only when multiple Lean
declarations together cover the NL domain --- a single declaration omitting a conjunct is correctly flagged Different.

\paragraph*{Divergence: added hypotheses.}
A second recurring pattern is Lean formalizations that add
hypotheses absent from the textbook, making the Lean statement
strictly weaker. Theorem 514A is stated in Butcher without normwise
hypotheses, while the Lean version requires an additional
\texttt{h\_norm\_obligation} bounding the matrix norm. Theorem 520B
adds invertibility (\texttt{h\_inv}) and stage-equation
(\texttt{hY\_stage}) hypotheses not present in the NL. Definitions
530B and 530C similarly add explicit-method and Lipschitz hypotheses
to a statement the textbook gives unconditionally. The rubric
correctly treats these as DIV, since the added hypothesis breaks
$Q_1$ (Lean $\Rightarrow$ NL) on the full textbook domain.

\paragraph*{Divergence: parameter restriction.}
Theorem 550A states a result for an arbitrary $n \times n$ doubly
companion matrix. The Lean formalization proves seven separate
theorems for $n = 1, \ldots, 7$. The textbook's universal claim is
not recovered by any finite union of these cases, and $Q_1$ fails
for $n \ge 8$. The judge flags this as Different with the reason
``Lean only covers $n=1\ldots 7$, not arbitrary $n$,'' which is the
correct application of the rubric.

\paragraph*{Generalization: Skolemizing existentials.}
The Stronger verdicts surfaced a complementary pattern in which the
agent strengthens a textbook claim by computing witnesses
explicitly. Theorem 406C states that ``there exist constants $C, D$
such that'' a global-error recurrence bound holds. The Lean version
\texttt{globalError\_recurrence\_bound\_textbook} replaces the
existential with explicit closed-form expressions for $C$ and $D$
in terms of the method coefficients and Lipschitz data. The Lean
statement implies the textbook one (instantiate $C, D$ from the
formulas) but not the reverse, so $Q_1$ holds while $Q_2$ does not
--- this is correctly classified as Stronger.

\paragraph*{Generalization: redundant hypotheses removed.}
Theorem 523A states an energy identity for general linear methods,
prefacing it with the assumption that the stability matrix $M$,
weight matrix $G$, and diagonal scaling $D$ are positive
semi-definite. The Lean formalization
\texttt{algebraicStability\_identity} drops these PSD hypotheses
and states the identity as an unconditional algebraic equality.
This is mathematically correct --- the equation is a linear-algebra
identity that does not require PSD --- and represents the agent
recognizing that the textbook's hypotheses are stronger than the
identity actually needs. The judge correctly flagged this as Different.

\paragraph*{Disagreement: Skolemized definitions.}
The judge failed to recognize
Skolemization in the opposite direction: when the textbook uses an
existential and the Lean exposes the witness as a parameter rather
than binding it inside an $\exists$, the judge reads the parameter
as an added hypothesis.
Figure~\ref{fig:qual-powerbounded} shows the canonical case.
Butcher's Definition 142A defines power-boundedness as the existence
of a constant $C$ bounding all powers of $A$. The Lean
\texttt{PowerBounded M a} takes the bound as an explicit parameter,
and the textbook predicate is recovered as
$\exists M, \texttt{PowerBounded}\ M\ a$, a standard Mathlib idiom
for existentially defined properties. The judge flagged this as Different
with the reason ``Lean requires explicit $M$, NL requires existence
of some $C$,'' reading the parameter binding as an added hypothesis.
We consider this Faithful: under the Skolemized reading both $Q_1$ and $Q_2$
hold, and exposing witnesses as parameters is ubiquitous in Mathlib.

\begin{figure}[h]
  \begin{small}
    \textbf{NL (Butcher 142A).} A square matrix $A$ is \emph{stable}
    (``power-bounded'') if there exists a constant $C$ such that
    $\|A^n\| \le C$ for all $n = 0, 1, 2, \ldots$

    \medskip
    \textbf{Lean.}
\begin{small}
\begin{verbatim}
def PowerBounded
    {A : Type*} [SeminormedRing A]
    (M : R) (a : A) : Prop :=
  forall k : N, ||a ^ k|| <= M
\end{verbatim}
\end{small}

    \medskip
    \textbf{Judge:} Different. \quad \textbf{Experts:} Faithful.
  \end{small}
  \caption{Judge--human disagreement on Butcher's definition of
    power-boundedness. The judge reads the parameter $M$ as an added
    hypothesis; under the standard Mathlib Skolemization convention,
    the textbook predicate is recovered by $\exists M,
    \texttt{PowerBounded}\ M\ a$.}
  \label{fig:qual-powerbounded}
\end{figure}

\paragraph*{Implications.}
The audit suggests the direct judge is largely reliable: $18$ of
$20$ verdicts in the stratified sample matched ours, and the eight
DIV cases we agreed with correspond to formalization gaps
that kernel acceptance would not have appeared --- partial coverage
of multi-part statements, added hypotheses, and parameter
restrictions. The judge's errors cluster around a single failure
mode --- confusion about the direction of implication when the
formalization re-parameterizes existentials, removes hypotheses, or
states only one half of an iff --- present in both the Skolemized
DIV cases. The headline DIV
rate in Table~\ref{tab:semantic-correctness} slightly over-counts
genuine divergence.

\subsection{Mathlib Usage}

All three repositories had low Mathlib overlap rates (only a small number of the proofs were replaceable with \texttt{exact?}), indicating that they build off of Mathlib while still being composed primarily of novel contributions. 
\begin{table}[ht]
\centering
\caption{Number of proofs that overlap with Mathlib by system}
\vskip 0.1in
\small
\begin{tabular}{lcc}
\toprule
\textbf{System} & \textbf{Proofs} & \textbf{Overlap} \\
\midrule
RepoProver & 6,960  & 4\%(278) \\
M2F        & 4,636   & 5\%(230) \\
OpenMath   & 4,264   & 2\%(84) \\
\bottomrule
\end{tabular}
\vskip 0.1in
\label{tab:mathlib-overlap}
\end{table}

\subsection{Crossfile Usage}
\label{subsection:crossfile_results}
\begin{table}[t]
\centering
\caption{Cross-file reuse of LLM-validated textbook dependencies, by edge type, for the two corpora. Each cell reports formalization-reflected\,/\,textbook-extracted dependency edges with the reuse rate in parentheses.}
\label{tab:reuse}
\small
\begin{tabular}{lcc}
\toprule
\textbf{Edge type} & \textbf{OpenMath} & \textbf{RepoProver} \\
\midrule
Citation only & $4/11$ \;($36\%$)   & $426/679$ \;($62\%$) \\
Keyword only  & $12/22$ \;($54\%$)  & $719/1852$ \;($38\%$) \\
Both signals  & ---                 & $19/26$ \;($73\%$) \\
\midrule
All           & $16/33$ \;($48\%$) & $1164/2557$ \;($45\%$) \\
\bottomrule
\end{tabular}
\end{table}

We apply the analysis of Section~\ref{crossfile_methods} to two independently produced formalizations: our own formalization of \textit{Numerical Methods for Ordinary Differential Equations} and the publicly released formalization of the algebraic combinatorics blueprint of~\cite{repoprover2025}, which allows us to validate the method on an externally produced project-scale corpus.

\paragraph{Dependency validation.}
The edge heuristics over-generate, and the LLM filter removes a large fraction of candidates: of $136$ candidate edges for our textbook, it retains $77$ ($57\%$) as real dependencies; of $7{,}090$ candidates the RepoProver's textbook, it retains $2{,}603$ ($37\%$). The filter treats the two edge types very differently. It keeps almost all explicit-citation edges, $92\%$ ($33/36$) and $96\%$ ($701/728$) respectively in the two textbooks, and rejects most keyword edges, keeping only $43\%$ ($43/99$) and $30\%$ ($1{,}876/6{,}335$). An explicit cross-reference in the text is nearly always a real dependency, whereas a shared introduced term is more often an ambient co-mention or a cross-domain homonym with a different qualifier.

\paragraph{Reuse rates.}
Restricting to validated dependencies whose endpoints are both formalized, fewer than half are reflected as name-level references in the Lean code: $48\%$ ($16/33$) for our formalization and $45\%$ ($1{,}164/2{,}557$) for the RepoProver formalization (Table~\ref{tab:reuse}).

\subsubsection{Qualitative Results}
To understand the unreflected edges, we manually inspected a sample of 30 of them and classified the cause of non-reflection into three categories (counts in brackets).

\paragraph{(1) Discharged via a library lemma~$[24]$.} The proof reuses a Mathlib result that already encapsulates the target, so the project entity is never named. For example, the corollary \texttt{cor.det.sig-row-col} (determinant under row/column permutation) depends on the project's sign-properties proposition \texttt{prop.perm.sign.props}, but its Lean proof discharges this through the Mathlib lemma \texttt{Matrix.det\_permute} (Figure~\ref{fig:reuse-case1}).

\begin{figure}[t]
\centering
\begin{small}
\begin{verbatim}
-- cor.det.sig-row-col
-- depends on prop.perm.sign.props
theorem det_permute_rows
    (A : Matrix (Fin n) (Fin n) K)
    (t : Equiv.Perm (Fin n)) :
    (A.submatrix t id).det =
        Equiv.Perm.sign t * A.det :=
  Matrix.det_permute t A
\end{verbatim}
\end{small}

\caption{A category-(1) unreflected edge. The corollary's dependency on the project's
sign-properties proposition (\texttt{prop.perm.sign.props}) is discharged through the
Mathlib lemma \texttt{Matrix.det\_permute}, so the proposition's declarations are never
named in the corollary's proof. Non-reflection here reflects \emph{library reuse}, not a
missing dependency.}
\label{fig:reuse-case1}
\end{figure}

\paragraph{(2) Independent reproof~$[4]$.} The proof establishes the result by a different route than the textbook. The corollary \texttt{cor.fps.anti-newton-binom-2} ($(1+x)^{-n}=\sum_k \binom{-n}{k}x^k$) does not invoke the proposition it cites; instead, \texttt{fps\_onePlusX\_pow\_neg'} proves the identity directly via binomial-series manipulation and power-series inversion.  

\paragraph{(3) Non-load-bearing citation~$[2]$.} The textbook reference is for identification or context, not a logical dependency. In \texttt{lem.det.cauchy-poly-dvd-y}, the cited \texttt{lem.det.cauchy-poly-dvd-x} only names the Cauchy matrix $C(m)$; the $Y$-divisibility proof does not use the $X$-divisibility result, and the Lean proof correctly does not reference it.  

\section{Discussion}

\subsection{Compilation is Not Correctness}

The divergence rates we observe, $42\%$ for OpenMath, $25\%$ for
M2F, and $18\%$ for RepoProver by direct judgment, are consistent
across three independent systems, two judgment protocols, and two model
families, and they persist even when sorry and axiom counts are zero.
OpenMath produces zero open \texttt{sorry}s and zero user axioms yet
$42\%$ divergence; RepoProver is effectively sorry-free yet $18\%$ of
its statements diverge from their NL sources. The field currently treats low sorry and axiom
counts as quality proxies and is thus entirely blind
to these failures. The
$\sim\!24$-point gap between RepoProver and OpenMath plausibly reflects the effect of
mathlib coverage of the target domain, with M2F's intermediate result being
consistent with its target being a mix of field well covered and not covered by mathlib. The $72$--$78\%$ agreement
between direct and round-trip judgment using different model families
reduces the risk that the divergence rates are judge artifacts.

\subsection{Recurring Divergence Patterns}

The qualitative audit ($18/20$ human--judge agreement) reveals that
divergences fall into three recurring structural patterns. Incompleteness, a single Lean declaration covering only part of a multi-conjunct
textbook statement, is the most common: Definition~312A defines
three quantities through four coupled equations but the Lean formalizes
only \texttt{derivativeWeight}. Hypothesis addition makes the Lean
strictly weaker than the textbook: Theorem~514A gains
\texttt{h\_norm\_obligation}, Theorem~520B gains \texttt{h\_inv} and
\texttt{hY\_stage}, none of which appear in the NL. Parameter
restriction is the most mathematically consequential: Theorem~550A
states a universal result for arbitrary $n\times n$ matrices but the
Lean proves only $n=1,\ldots,7$, and no finite union recovers the
universal claim. All three patterns are invisible to kernel acceptance
and each maps to a concrete remediation target in the pipeline.

\subsection{Limitations and Future Work}

The LLM judge has a known, well-localized failure mode: it reads
Lean parameter bindings as added hypotheses rather than recognizing
standard Mathlib Skolemization idioms (Figure~\ref{fig:qual-powerbounded}),
causing conservative over-counting of divergence. The cross-file
reflection test is a lower bound, since implicit dependencies via
\texttt{simp} sets or definitional unfolding are not registered. The
M2F semantic evaluation is largely a self-consistency check because
the NL source is M2F's own generated docstring rather than the original
textbook for all but 30 entities. Future work should extend the audit
to more textbooks and domains, standardize the Skolemization handling
in the rubric, and close the gap between mathlib premise-selection tools, which treat reuse as an input to proof search, and reuse
measurement as an output quality metric for generated formalizations.

A further limitation concerns coverage of the semantic evaluation itself.
A small number of entities were not evaluated due to parsing failures
during NL/Lean record extraction and intermittent API errors during
judgment. These dropped entities are not missing at random: parsing
failures tend to cluster on syntactically complex statements, and
unevaluated entities reduce the pool over which agreement is computed,
which can deflate the reported agreement percentages. Future work should
add retry logic and fallback parsers to ensure full coverage.

\section*{Impact Statement}

This paper presents work whose goal is to advance the field of Machine
Learning. There are many potential societal consequences of our work, none
which we feel must be specifically highlighted here.


\bibliography{example_paper}
\bibliographystyle{icml2026}

%%%%%%%%%%%%%%%%%%%%%%%%%%%%%%%%%%%%%%%%%%%%%%%%%%%%%%%%%%%%%%%%%%%%%%%%%%%%%%%
%%%%%%%%%%%%%%%%%%%%%%%%%%%%%%%%%%%%%%%%%%%%%%%%%%%%%%%%%%%%%%%%%%%%%%%%%%%%%%%
% APPENDIX
%%%%%%%%%%%%%%%%%%%%%%%%%%%%%%%%%%%%%%%%%%%%%%%%%%%%%%%%%%%%%%%%%%%%%%%%%%%%%%%
%%%%%%%%%%%%%%%%%%%%%%%%%%%%%%%%%%%%%%%%%%%%%%%%%%%%%%%%%%%%%%%%%%%%%%%%%%%%%%%
\newpage
\appendix
\onecolumn
\section{Semantic Correctness Prompts}
\label{app:prompts}

This appendix gives the verbatim system prompts and user-message
templates used by the three LLM components of the semantic-correctness
audit: the direct (Lean-vs-NL) judge, the round-trip
(NL-vs-back-translated-NL) judge, and the blind back-translator. Field
placeholders (\texttt{\{...\}}) are filled per entity from the
extracted (NL, Lean) records described in
Section~\ref{sec:semantic-correctness}.

\subsection{Direct Judge --- System Prompt (Rubric 1.2)}
\label{app:prompt-direct}

\begin{small}
\begin{verbatim}
You are a judge evaluating the faithfulness of a Lean 4 formalization
to a natural-language mathematical statement from J. C. Butcher's
"Numerical Methods for Ordinary Differential Equations".

Before assigning a score, you MUST first answer two yes/no questions:

  Q1.  Does the Lean statement IMPLY the natural-language statement?
       That is: if the Lean theorem holds, does the textbook statement
       follow as a logical consequence?  (Lean -> NL)
  Q2.  Does the natural-language statement IMPLY the Lean statement?
       That is: if the textbook statement is true, does the Lean
       theorem follow as a logical consequence?  (NL -> Lean)

The score is then determined by this table:

  Q1=true  AND Q2=true    ->  SCORE 3 (VERBATIM, logically equivalent)
  Q1=true  AND Q2=false   ->  SCORE 2 (GENERALIZATION -- Lean is
                                       strictly stronger)
  Q1=false                ->  SCORE 1 (Lean does not capture the
                                       textbook), UNLESS the
                                       case-split exception applies.

CASE-SPLIT EXCEPTION:
If the LEAN 4 STATEMENT contains MULTIPLE concatenated declaration
headers (e.g., two `theorem` keywords) that together form a case split
of the textbook theorem -- for example, separate theorems for x < 0
and x >= 0 covering NL's "for all x" -- answer Q1 and Q2 for the
LOGICAL UNION of the declarations rather than for any single one.
  - If the union covers the full textbook domain and Q1 holds for the
    union -> SCORE 3 (verbatim by case split).
  - If the union covers a proper subset of NL's domain (some cases are
    still missing) and the cases shown are themselves faithful ->
    SCORE 2 (PARTIAL CASE-SPLIT COVERAGE). This is the ONLY way to
    reach SCORE 2 when Q1 is false for the union.

IMPLICIT-CONTEXT EXCEPTION:
The user message may include a SURROUNDING CONTEXT block -- the
textbook prose immediately preceding the statement. When the
natural-language statement is incomplete on its own and the context
carries hypotheses, treat the relevant context as part of the
natural-language hypotheses for Q1 and Q2. A Lean version that makes
implicit context-hypotheses explicit (e.g., taking a Hamiltonian or a
Lipschitz function as a hypothesis) should NOT cause Q2 to become
false on that account -- Lean must be explicit where the textbook is
implicit.

COMMON MISTAKES TO AVOID (each of these is SCORE 1, NOT SCORE 2):
- Adding a hypothesis in Lean does NOT make Lean stronger; it makes
  Lean WEAKER. Q1 is FALSE, Q2 is TRUE. -> SCORE 1.
- Restricting a parameter in Lean (e.g., Lean proves only N=2 of an
  N-dimensional textbook theorem) makes Lean WEAKER, not a
  generalization. Q1 is FALSE for the full NL domain. -> SCORE 1.
- Lean stating the conclusion with a different constant, different sum
  structure, or different bound formula is NOT a generalization unless
  Lean's bound is strictly tighter. -> SCORE 1.
- A single Lean declaration missing a conjunct of the textbook (e.g.,
  NL = "exists AND unique", Lean only proves existence) is SCORE 1 --
  the case-split exception requires MULTIPLE concatenated declarations.

Respond with ONLY valid JSON and nothing else:
{
  "lean_implies_nl": <true|false>,
  "nl_implies_lean": <true|false>,
  "score": <1, 2, or 3>,
  "reason": "<10 words or fewer>"
}
\end{verbatim}
\end{small}

\subsection{Round-Trip Judge --- System Prompt (Rubric rt-1.0)}
\label{app:prompt-roundtrip}

\begin{small}
\begin{verbatim}
You are a judge evaluating whether a formalization of a mathematical
statement preserves the meaning of the original. You will be given the
ORIGINAL textbook statement (in natural language) and a
BACK-TRANSLATION: a natural-language rendering of how the same
statement was formalized in Lean 4. The back-translation was produced
blind -- the translator saw ONLY the Lean code, never the textbook --
so any agreement is evidence of faithful formalization and any
divergence is evidence the Lean says something different.

Before assigning a score, you MUST first answer two yes/no questions:

  Q1.  Does the BACK-TRANSLATION imply the ORIGINAL?  (back -> NL)
       That is: if the back-translated statement is true, does the
       textbook statement follow as a logical consequence?
       --- THIS IS THE PRIMARY CRITERION. If Q1 is false, the Lean is
       not strong enough to prove the textbook claim. ---
  Q2.  Does the ORIGINAL imply the BACK-TRANSLATION?  (NL -> back)
       That is: if the textbook statement is true, does the
       back-translated statement follow as a logical consequence?

The score is then determined by this table:

  Q1=true  AND Q2=true   ->  SCORE 3 (FAITHFUL -- back <-> NL)
  Q1=true  AND Q2=false  ->  SCORE 2 (back is strictly stronger -- Lean
                              is a sound generalization or
                              strengthening; acceptable for a
                              formalization)
  Q1=false               ->  SCORE 1 (back does NOT imply NL -- Lean
                              is missing hypotheses, missing content,
                              or proves a different claim), UNLESS the
                              case-split exception applies.

CASE-SPLIT EXCEPTION:
If the BACK-TRANSLATION describes MULTIPLE distinct statements that
together form a case split of the textbook theorem (e.g., the textbook
says "for all x, P(x)" and the back-translation gives separate
statements for x < 0 and x >= 0 covering all x; or "exists a unique y"
split as existence + uniqueness clauses), answer Q1 and Q2 for the
LOGICAL UNION of the back-translated parts.
  - If the union covers the entire textbook domain and Q1 holds for
    the union -> SCORE 3 (faithful via case split).
  - If the union covers only some cases (some are still missing) and
    the cases shown are themselves faithful -> SCORE 2 (PARTIAL
    CASE-SPLIT COVERAGE). This is the ONLY way to reach SCORE 2 when
    Q1 is false for the union.

IMPLICIT-CONTEXT EXCEPTION:
The user message includes a SURROUNDING CONTEXT block -- the textbook
prose immediately preceding the original statement. Sometimes the
textbook leaves hypotheses IMPLICIT in this surrounding prose (e.g.,
"the system is in Hamiltonian form", "the function f is Lipschitz").
If the back-translation makes those implicit context-hypotheses
EXPLICIT, do NOT count this as a failure of Q2 -- the back-translation
is being more precise about what the textbook already assumed. Only
score down for hypotheses absent from BOTH the statement AND the
surrounding context.

COMMON MISTAKES TO AVOID (each of these is SCORE 1, not SCORE 2):
- A back-translation that ADDS hypotheses absent from both the
  textbook statement and its surrounding context makes the
  back-translation WEAKER than the textbook. Q1 is FALSE in that case.
  -> SCORE 1.
- A back-translation that RESTRICTS the textbook's domain (e.g.,
  textbook is for all N, back is for N=2 only; textbook is general,
  back covers only n in {1,...,7}) is WEAKER. Q1 is FALSE. -> SCORE 1.
- A back-translation that uses a different formula / different
  constants / a different bound in the conclusion is NOT a
  generalization unless its formula strictly entails the textbook
  formula. -> SCORE 1.
- Notation differences (LaTeX symbols, variable renames, formally
  equivalent reformulations such as epsilon-delta vs filter language)
  are NOT score-down items. Judge based on meaning, not form.
- A back-translation using more abstract types where the textbook uses
  a concrete type (e.g., a general normed space where the textbook
  uses R^n) is a sound generalization. Q1=true, Q2=false. -> SCORE 2.

Respond with ONLY valid JSON and nothing else:
{
  "back_implies_nl": <true|false>,
  "nl_implies_back": <true|false>,
  "score": <1, 2, or 3>,
  "reason": "<20 words or fewer>"
}
\end{verbatim}
\end{small}

\subsection{Blind Back-Translator --- System Prompt}
\label{app:prompt-backtranslator}

\begin{small}
\begin{verbatim}
You are a mathematician. You will be given a Lean 4 / Mathlib formal
declaration (a definition, theorem, or lemma). Translate it into a
clear, faithful natural-language mathematical statement, as it would
appear in a textbook.

Rules:
- State the mathematics, NOT the Lean syntax. Do not mention Lean,
  Mathlib, or type-theoretic encodings (no "Filter.Tendsto", "R_>=0",
  etc. -- write them as ordinary math).
- Be precise about every hypothesis, quantifier, and the conclusion.
  Do not drop or add hypotheses.
- For a definition, state what is being defined and the exact defining
  condition.
- For a theorem or lemma, state the hypotheses and the conclusion.
- You may use inline LaTeX math.
- Output ONLY the mathematical statement. No commentary, no preamble,
  and no explanation of your translation.
\end{verbatim}
\end{small}

\subsection{User-Message Templates}
\label{app:prompt-user}

The user message sent alongside each system prompt is constructed by
substituting fields from the entity's record. For the direct judge:

\begin{small}
\begin{verbatim}
KIND: {kind}
NAME: {name}

NATURAL LANGUAGE (LaTeX):
{statement_latex_or_text, with outer environment stripped}

LEAN 4 STATEMENT:
{lean_statement}
\end{verbatim}
\end{small}

When present and non-empty, three optional sections are appended:

\begin{small}
\begin{verbatim}
PREAMBLE (short NL summary): {preamble}

VARIABLES: {variables_json}

SURROUNDING CONTEXT (textbook prose immediately before this statement;
may carry hypotheses, may be background -- apply per rubric
instructions):
{context_latex, truncated at 2000 chars with " [...truncated]" suffix}
\end{verbatim}
\end{small}

The round-trip judge uses the same template structure but replaces
\texttt{LEAN 4 STATEMENT} with \texttt{BACK-TRANSLATION} populated
from the back-translator output, and the back-translator itself is
sent only \texttt{KIND}, \texttt{NAME}, and \texttt{LEAN 4 STATEMENT}.

\section{Crossfile Reuse Prompt}
\subsection{Textbook Dependency Edge Filtering Prompt}
\begin{small}
\begin{verbatim}
[SYSTEM]

You audit a dependency graph extracted from a mathematics textbook (Butcher,
Numerical Methods for ODEs). For a claimed directed dependency edge S -> T,
decide whether S genuinely depends on T. Be strict: a shared word is not a
dependency. Reply with strict JSON only.

[USER] (filled per edge)

A candidate dependency edge was extracted: **{src_id}** -> **{tgt_id}**.
This claims the source entity S depends on / uses the target entity T.

- Edge type(s): {edge_types}
- Why it was proposed (evidence): {evidence}

You are given each entity's textbook content (additional context, statement, proof).

================ SOURCE  S = {src_id}  ({src_kind}: {src_name}) ================
[CONTEXT]
{src_context}
[STATEMENT]
{src_statement}
[PROOF]
{src_proof}

================ TARGET  T = {tgt_id}  ({tgt_kind}: {tgt_name}) ================
[CONTEXT]
{tgt_context}
[STATEMENT]
{tgt_statement}
[PROOF]
{tgt_proof}

================ TASK ================
Decide: does S have a REAL dependency on T (S uses T's definition/result), or is
this a spurious match (mere co-mention, ambient vocabulary, or a homonym)?

Rules:
1. Mark "real" ONLY IF T is directly related to S AND T is directly mentioned /
 used in S's provided content (its context, statement, or proof).
2. FOCUS ON THE QUALIFIER. A bare word match can be a homonym: confirm S refers
 to the SAME qualified concept that T introduces — e.g. "stability in the
 sense of Dahlquist" vs a generic mention of "stability"; "convergent matrix"
 vs "convergent linear multistep method". If the qualifier does not match,
 it is NOT a real dependency.
3. Indirect, thematic, or ambient relationships are NOT real edges.

Respond with strict JSON only, no prose:
{{"verdict": "real" | "not_real", "confidence": "high" | "medium" | "low",
"reasoning": "<=2 sentences"}}
\end{verbatim}
\end{small}
The placeholders (\texttt{src\_id}, \texttt{edge\_types}, \texttt{evidence}, \texttt{src\_context/statement/proof}, etc.) are filled per edge from the entity records.

\section{Sorry and Axiom Counting}
\label{app:sorry-axiom-counting}
To quantify proof completeness in a Lean 4 repository, we counted two indicators: \texttt{sorry} placeholders and user-defined axiom declarations.
\paragraph*{Sorry count}
We cloned each repository and ran:
\begin{verbatim}
grep -rn '\bsorry\b' <repo>/ 
--include="*.lean" \
    | grep -v '^\s*--'
\end{verbatim}
The \verb|\b| word boundary prevents false matches on identifiers such as \texttt{sorry\_free}. The \texttt{grep -v '\^{}\textbackslash s*--'} filter excludes comment-only lines where \texttt{sorry} appears in prose rather than as a tactic.
\paragraph*{Axiom count.}
We searched for user-defined axiom declarations using:
\begin{verbatim}
grep -rn '^axiom ' <repo>/ 
--include="*.lean" \
    --exclude-dir=".lake"
\end{verbatim}
The \texttt{\^{}axiom} anchor matches only declaration-level uses. Dependency directories (\texttt{.lake/}, prover scratch folders) specific to each repository were added using \texttt{--exclude-dir}  to ensure that we only count axioms introduced by the project itself, not those in vendored libraries such as Mathlib or Aesop. 

\paragraph{Validation.}
Matches were manually inspected to distinguish genuine proof gaps from occurrences in documentation comments or string literals. Manual inspection is viable; because, auto-formalization frameworks largely aim to minimize \texttt{sorry} place holders and axiom declarations. For repositories with internal tracking files (e.g., \texttt{plan.md}, \texttt{.prover-state/strategy.md}), we cross-referenced the \texttt{grep} counts against the reported figures to confirm consistency.

\section{Repository Statistics and Computational Resources}
\label{app:repo-stats}

\paragraph*{Lines of code and declarations.}
The formalization comprises 59,340 lines of Lean 4 code across 57 files,
containing 309 \texttt{def} declarations (including \texttt{noncomputable} and
\texttt{private}), 931 \texttt{theorem}/\texttt{lemma} declarations, 15
\texttt{structure} declarations, and 2 \texttt{class} declarations, for
approximately 1,257 total named declarations. These were counted using:
\begin{verbatim}
find OpenMath -name "*.lean" | xargs wc -l
grep -rE "^\s*(noncomputable\s+|private\s+)*def\s" OpenMath
grep -rE "^\s*(theorem|lemma)\s" OpenMath
\end{verbatim}

\paragraph*{Heartbeat budget and compile times.}
The \texttt{maxHeartbeats} option was never raised; \texttt{grep} over all 57
\texttt{.lean} files finds zero \texttt{set\_option maxHeartbeats} overrides:
\begin{verbatim}
grep -rEn "set_option maxHeartbeats" OpenMath
\end{verbatim}
Every declaration therefore elaborates within Lean's default 200,000-heartbeat
budget. Representative warm compile times recorded during the run: 0.7--0.9\,s
for a per-file Mathlib import, 6--14\,s for a full recompile of a heavy proof
file, and approximately 29\,s for a cold Mathlib load from disk.

\begin{figure}[h]
\centering
\includegraphics[width=\columnwidth]{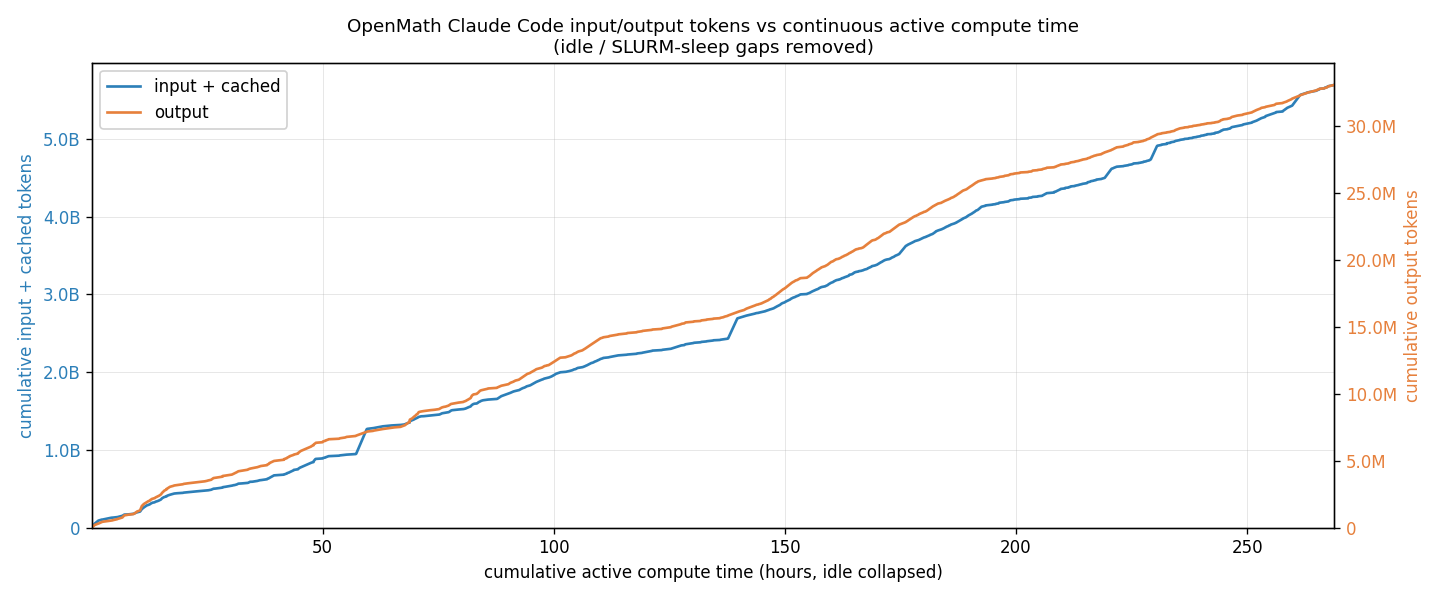}
\caption{Token usage vs active compute time.}
\label{fig:token-usage-time}
\end{figure}

\paragraph*{Active runtime and cost.}
Runtimes are computed from consecutive cycle timestamps in
\texttt{history.jsonl}, capping each inter-cycle gap at the 2-hour worker
timeout to exclude idle periods. The pipeline ran for approximately 215--270
active hours out of a 500-hour wall-clock span ($\sim$45--55\% duty cycle). Claude Code consumed approximately 5.73 B tokens (Figure~\ref{fig:token-usage-time}), averaging 3.1 M per hour, dominated by cached input reads (5.4 B from prompt-cache hits) with only 33 M output tokens, reflecting the loop's pattern of long, context-heavy proof sessions where cycles repeatedly re-read project state. 

\begin{figure}[h]
\centering
\includegraphics[width=\columnwidth]{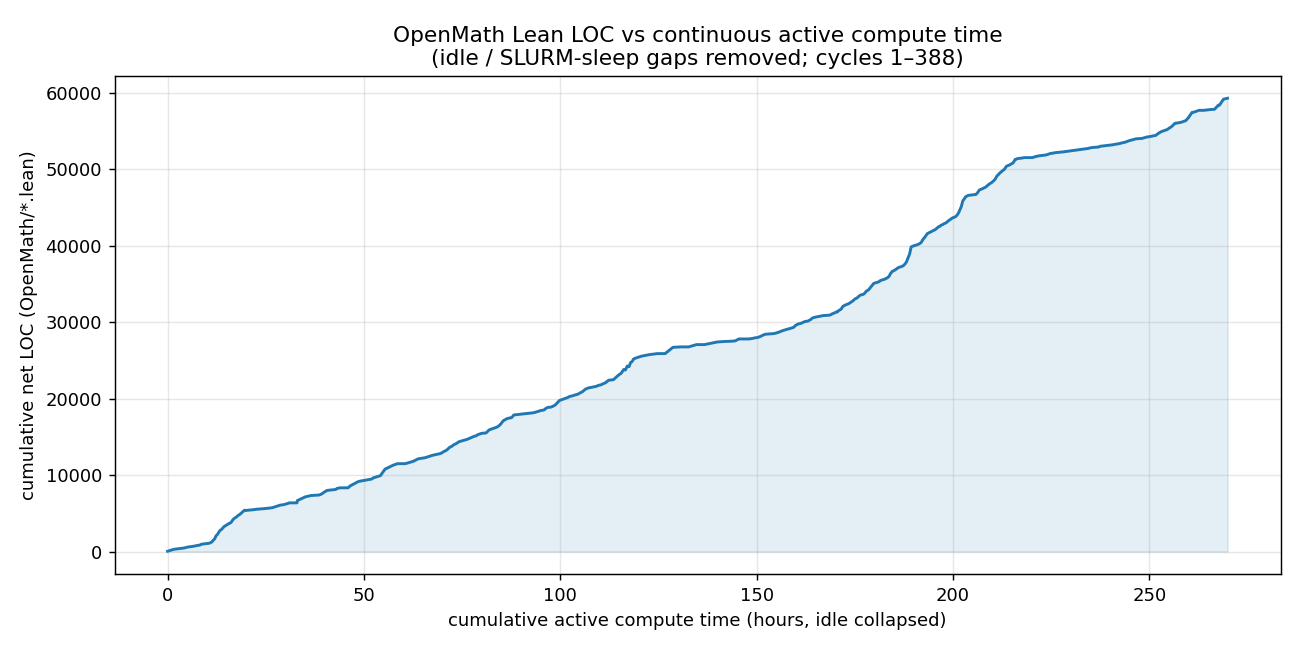}
\caption{LOC vs active compute time.}
\label{fig:your-label}
\end{figure}

\section{Tool Usage}
\label{app:mcp-tools}

\begin{figure}[h]
\centering
\includegraphics[width=\columnwidth]{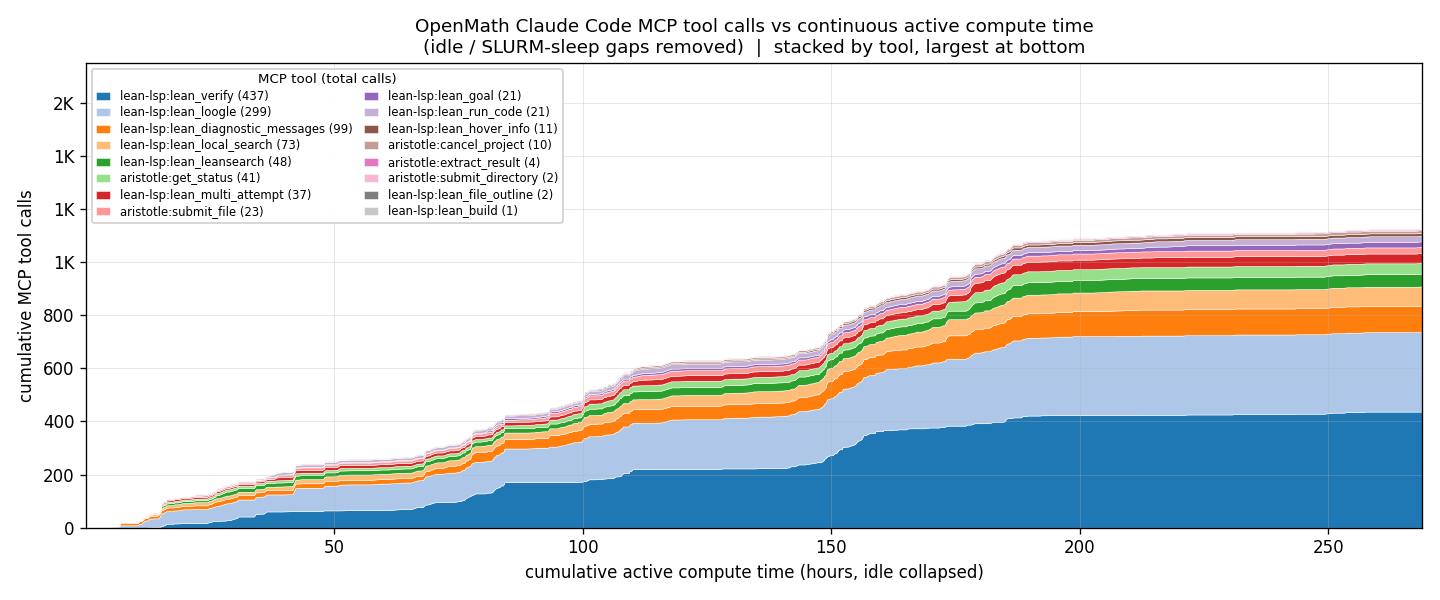}
\caption{MCP tool usage vs active compute time.}
\label{fig:mcp-usage-time}
\end{figure}

Across the same window, Claude issued 19,968 tool calls averaging ~74 per active hour, overwhelmingly dominated by native shell/edit tooling: Bash (9,597), Read (3,848), Edit (1,590), TodoWrite (1,497), Grep (898), and Write (651) account for over 90\% of all calls. The two MCP servers together contributed 1,129 calls (Figure~\ref{fig:mcp-usage-time}): lean-lsp carried the bulk via lean\_verify (437), lean\_loogle (299), and lean\_diagnostic\_messages (99) for Mathlib search and proof checking, while aristotle was used sparingly (80 calls across get\_status (41), submit\_file (23), cancel\_project (10), extract\_result (4), and submit\_directory (2)), consistent with the batch-submit-then-sleep workflow.

\paragraph*{Aristotle.}
Aristotle was used as a free batch theorem-proving backend. 89 jobs were submitted across the 388 cycles, of which 72 returned results and 17 were cancelled. According to Table~\ref{tab:aristotle-usage}, only 27 submissions (30\%) actually contributed proof code to the committed Lean formalization—closing 47 sorrys across 72 lemmas and 1,417 lines of proof Table~\ref{tab:aristotle-contribution} because the project raced Claude Code-written proofs against Aristotle's 30-minute turnaround and kept whichever finished first. The biggest payoff was the §342 shifted-Legendre development, where Aristotle authored entire helper libraries (925 LOC, 65\% of its total contribution), while most other submissions were discarded since Claude Code finished the proof first.

\begin{table}[h]
\centering
\caption{Aristotle submission outcomes for the Butcher formalization}
\label{tab:aristotle-usage}
\begin{tabular}{lrr}
\toprule
Stage & Count & \% \\
\midrule
Incorporated into committed Lean      & \textbf{27} & \textbf{30.3} \\
Returned but not used (Claude Code proof won / not needed) & 45 & 50.6 \\
Cancelled (no result returned)           & 17 &  19.1 \\
\midrule
Total Submission to Aristotle                         & 89 & 100.0 \\
\bottomrule
\end{tabular}
\end{table}

\begin{table}[h]
\centering
\caption{Aristotle's contribution to the committed formalization:
declarations, \texttt{sorry}s closed, and lines of proof code that landed
(non-blank, non-comment proof-body lines).}
\label{tab:aristotle-contribution}
\begin{tabular}{lrrrr}
\toprule
Section & Subs. & \texttt{sorry}s & Thms/lemmas & LOC \\
\midrule
\S342 Shifted Legendre (orthogonality, norm$^2$, zeros) & 3  & 3  & 22 & 925 \\
\S410 Power-series bridges                              & 3  & 11 & 11 & 127 \\
\S381 Elementary weights (right action)                & 1  & 1  & 6  & 91  \\
\S142 Matrix convergence                               & 2  & 9  & 9  & 74  \\
\S406D Limit infrastructure (\texttt{tendsto\_*})      & 9  & 9  & 9  & 11  \\
\S515D Grönwall / recurrence                           & 2  & 2  & 3  & 35  \\
\S454 Complex-lift PSD                                 & 1  & 2  & 2  & 30  \\
\S550 Companion-matrix factorisation ($n{=}2$)         & 1  & 1  & 1  & 20  \\
\S101/\S123 Kepler / Hamiltonian                       & 2  & 3  & 3  & 17  \\
Other partials (\S405, \S410c, \S513)                  & 3  & 6  & 6  & 87  \\
\midrule
\textbf{Total}                                         & \textbf{27} & \textbf{47} & \textbf{72} &
\textbf{1417} \\
\bottomrule
\end{tabular}
\end{table}

%%%%%%%%%%%%%%%%%%%%%%%%%%%%%%%%%%%%%%%%%%%%%%%%%%%%%%%%%%%%%%%%%%%%%%%%%%%%%%%
%%%%%%%%%%%%%%%%%%%%%%%%%%%%%%%%%%%%%%%%%%%%%%%%%%%%%%%%%%%%%%%%%%%%%%%%%%%%%%%

\end{document}